\documentclass[conference]{IEEEtran}

\ifCLASSOPTIONcompsoc
  \usepackage[nocompress]{cite}
\else
  \usepackage{cite}
\fi

\usepackage[pdftex]{graphicx}

\usepackage[float]{}
\usepackage{algorithm}
\usepackage{algorithmicx}
\usepackage[noend]{algpseudocode}

\usepackage{algorithm}
\usepackage{algpseudocode}
\usepackage{textcomp}
\usepackage[english]{babel}
\usepackage[utf8x]{inputenc}
\usepackage{amsmath}
\usepackage{amssymb}
\usepackage{graphicx}
\usepackage[export]{adjustbox}
\usepackage{array,multirow}
\usepackage[table]{xcolor}
\usepackage{listings}
\usepackage{cleveref}
\usepackage{url}
\usepackage{arydshln}
\usepackage{microtype}
\graphicspath{ {figures/} }
\makeatletter
\DeclareTextCommandDefault{\textleftarrow}{\mbox{$\m@th\leftarrow$}}
\makeatother

\newcommand{\call}[1]{\textit{#1}}

\DeclareMathOperator*{\argmax}{argmax}

\begin{document}

\title{Generating Diverse and Competitive Play-Styles\\ for Strategy Games}

\IEEEoverridecommandlockouts
\IEEEpubid{\begin{minipage}{\textwidth}\ \\[12pt]
978-1-6654-3886-5/21/\$31.00 \copyright 2021 IEEE
\end{minipage}}

\author{
\IEEEauthorblockN{Diego Perez-Liebana, Cristina Guerrero-Romero, \\Alexander Dockhorn, Linjie Xu, Jorge Hurtado}
\textit{Game AI Group}\\
\IEEEauthorblockA{\textit{Queen Mary University of London, UK}}
\and
\IEEEauthorblockN{Dominik Jeurissen\\}
\IEEEauthorblockA{
\textit{Games \& AI Group}\\
\textit{Maastricht University}\\
\textit{The Netherlands}}
}
\IEEEtitleabstractindextext{%
\begin{abstract}

Designing agents that are able to achieve different play-styles while maintaining a competitive level of play is a difficult task, especially for games for which the research community has not found super-human performance yet, like strategy games. These require the AI to deal with large action spaces, long-term planning and partial observability, among other well-known factors that make decision-making a hard problem. On top of this, achieving distinct play-styles using a general algorithm without reducing playing strength is not trivial. In this paper, we propose Portfolio Monte Carlo Tree Search with Progressive Unpruning for playing a turn-based strategy game (Tribes) and show how it can be parameterized so a quality-diversity algorithm (MAP-Elites) is used to achieve different play-styles while keeping a competitive level of play. Our results show that this algorithm is capable of achieving these goals even for an extensive collection of game levels beyond those used for training.
\end{abstract}
}

\maketitle

\IEEEdisplaynontitleabstractindextext

\section{Introduction} \label{sec:intro}

A large body of the current research on game playing AI agents is driven by the same performance metric: achieving a high win rate. From Go~\cite{silver2016mastering} to StarCraft~\cite{vinyals2019alphastar, ontanon2013survey}, including works on different general video-game systems like the Arcade Learning Environment~\cite{mnih2013playing} and the General Video Game AI framework~\cite{perez2019general}, the objective is to achieve the highest level of play possible. This is not different in complex strategy games, where the decision-making problem is far from trivial. Examples of these games are the already mentioned StarCraft, or other turn-based strategy games such as Blood Bowl~\cite{justesen2019blood} and Tribes~\cite{perez2020tribes}.
In this game genre, winning is a hard enough problem, but not the only one. When looking at games from the angle of the entertainment (an angle specially appealing to the games industry), there is a particular interest in creating AI opponents that are not only challenging, but also fun to play against. It is often the case that a game designer tries to achieve this by having games in which players can follow different strategies to achieve victory.

This paper tackles both issues at once: we study how a known algorithm (Monte Carlo Tree Search - MCTS~\cite{browne2014MCTSsurvey}) can be modified to incorporate different play-styles while being competitive. We first present an adaptation of MCTS that incorporates a portfolio of scripts coupled with Progressive Unpruning for managing the large action space of a turn-based strategy game (Tribes~\cite{perez2020tribes}). Although similar work exists about using a top-level search algorithm like MCTS with a portfolio of scripts~\cite{barriga2015puppet, churchill2015hierarchical}, our paper incorporates progressive unpruning and its parameterization to achieve different and competitive play-styles. The latter is done via a quality-diversity method (MAP-Elites~\cite{mouret2015illuminating}), which explores diverse play-styles by extracting different game-play traits, extending the evaluation to game levels that have not been used for training.

Section~\ref{sec:back} describes Tribes and the relevant algorithms; Section~\ref{sec:agent} explains our proposed Portfolio MCTS agent, and Section~\ref{sec:mapelites} the implementation of MAP-Elites. Section~\ref{sec:res} shows our experimental setting and results, and the paper finishes in Section~\ref{sec:end} with conclusions and ideas for future work.


\section{Background} \label{sec:back}

\subsection{Tribes}

Tribes~\cite{perez2020tribes} is an open source implementation of the award-wining mobile game ``The Battle of Polytopia'' (Midjiwan AB, 2016). It is a turn-based strategy game where two or more factions (or tribes) compete to be the last player standing (by capturing all the opponents' capital cities) or becoming the player with the highest score at turn $50$. Factions must master technology research, economy management and combat to win. The combination of these factors requires effective decision-making and allows for different strategies and play-styles.  



The game takes place in a two-dimensional grid of $N \times N$ cells, with each player starting with a capital city and a single unit. Each tile is of a particular terrain type%
, may hold a resource type 
or contain a city (owned by a player or a neutral village that can be captured). During the game, each player controls multiple units with different characteristics, recruitment costs, and abilities. They can be melee (warriors, riders, defenders, swordsmen and knights), ranged (archers and catapults) or special units (mind benders and superunits). 
Each city owned by a player gives control of the surrounding land, which permits the city to gather resources and construct buildings to increase the city's population. Increasing the population of a city will progressively allow the city to level up, which increments the number of \textit{stars} (the game's currency) the city provides at the start of each turn. This increments the tribe's capability to conduct research, construct more buildings and spawn combat units. Which buildings and units are available depend on which of the $24$ available technologies on the research tree have been completed so far by the tribe. 

Tribes is implemented in Java and provides an API for agents with access to a Forward Model (FM).
The framework includes different agents, among which
Monte Carlo Tree Search (MCTS;~\cite{browne2014MCTSsurvey}) 
and 
Rolling Horizon Evolutionary Algorithm (RHEA;~\cite{gaina2021rolling})
are relevant to and used in this study.

There are $21$ types of actions available in the game for each one of the units, cities and factions (see Table~\ref{tab:tabScripts} at the end of this paper). Each turn a player can execute many of these actions as long as they have enough stars to pay for them and there are available units to move. \cite{perez2020tribes} highlights the complexity of this game: in a typical game, the strongest agents in the framework (RHEA, RB and MCTS, in this order), show an average of $54.47$ possible actions to choose from at every decision point. The turn branching factor averages $10^7$ and $10^{23}$ (for early and end game, respectively), with winning bots normally reaching $10^{32}$ in the last $5$ turns of the game. The size of the game tree, for $2$ players and $50$ turns, is approximately $10^{1500}$. For a more complete definition of the game, the implemented agents, their relative playing performance and game complexity, the reader is referred to~\cite{perez2020tribes}.

\subsection{Statistical Forward Planning Agents for Strategy Games}

Implementing AI agents for strategy games is a tough challenge due to the unique properties of this genre. Rule-based AI agents usually implement a set of basic strategies to be followed for each of the game-playing components, e.g. fighting, research, and economy. Since those rule-based agents need to be specifically developed for each game, much research has been put into statistical forward planning agents which, given a forward model, should be able to play the game without further knowledge~\cite{Lelis2020}.
However, the complexities that strategy games bring still need to be addressed, such as huge branching factors or the needs for long-term planning.

The statistical forward planning algorithm MCTS has successfully been applied to many problem domains including strategy games such as Starcraft 1 \& 2~\cite{ontanon2013survey} and  Stratega~\cite{dockhorn2020stratega}.
MCTS is a tree search based approach which focuses on promising parts of the tree for further simulations.
To select tree nodes, MCTS uses Upper Confidence Bounds (UCB), a policy that balances between exploration and exploitation during the search.
While the use of UCB has been proven to converge to the optimal decision~\cite{browne2014MCTSsurvey}, the process can be slow in terms of required simulations.
Strategies such as progressive bias~\cite{chaslot2008}, hard pruning~\cite{hsu2020mctsPruning} and progressive widening~\cite{coulom2007} have shown effective in improving the algorithm's performance by incorporating domain knowledge into the search tree or focusing the search on a few promising nodes.
These are explained in \Cref{sec:pmcts}, where our agent is described.

Rolling Horizon Evolutionary Algorithms (RHEA) have shown effective in playing a wide range of games~\cite{gaina2021rolling}. 
RHEA agents create a population of possible action sequences and evolve those over the course of multiple generations.
Due to its known performance in other strategy games~\cite{dockhorn2020stratega}, we use the RHEA agent as a baseline in our evaluations.

To address the huge branching factor of strategy games, action abstraction methods can be used to reduce the number of available actions to a promising subset. Portfolio methods have shown to perform well in the context of strategy games~\cite{Dockhorn2021}.
In contrast to algorithms that search for the best action, portfolio methods reduce the search space to a set of scripts, with each one selecting an  action for a given game-state or unit. 
Portfolios have been used in conjunction with various search algorithms.
The simplest instance is Portfolio Greedy Search (PGS) which uses hill-climbing to optimize a set of script assignments~\cite{Churchill2013}.
While PGS iteratively optimizes the opponent's and the player's script assignment, Nested-Greedy Search (NGS)~\cite{moraes2018nested} evaluates the player's script assignment according to the best possible opponent's script assignment.
Further reductions of the search space have been achieved by Stratified Strategy Selection (SSS)~\cite{lelis2017stratified}.
Instead of assigning a script to a single unit, units are first grouped into types, then each type is assigned a script.

While portfolio algorithms are able to reduce the search space considerably, they may also ignore solutions that are not returned by any script.
Methods for asymmetric abstraction~\cite{moraes2018asymmetric} apply the abstraction only for certain parts of the game-state.
This principle has been used in the A3N algorithm~\cite{moraes2018action}, which groups units into unrestricted and restricted units.
While unrestricted units can be assigned all legal unit-actions only the actions returned by a set of scripts will be considered for restricted units.
In a survey by Lelis~\cite{Lelis2020}, the above portfolio methods have been compared and unified in the General Combinatorial Search for Exponential Action Spaces (GEX) algorithm.
Seeing how those different algorithms emerge as instances from a unifying framework allows to easily explore further variants of action abstracting search algorithms.



\subsection{MAP-Elites}

The previous algorithms focus on maximizing the agent's win-rate in strategy games. 
Next to improving the quality of an agent, we are interested in the optimization of diverse agents with unique play-styles.
The Multi-dimensional Archive of Phenotypic Elites (MAP-Elites)~\cite{mouret2015illuminating}) is a quality-diversity algorithm that has recently been used to create a diverse set of game-playing agents~\cite{canaan2019diverse} and game-play elements~\cite{khalifa2018talakat}.

The algorithm uses a behaviour-function that returns a vector representing the agent's behaviour, corresponding to a set of N phenotypical features.
This behaviour-space is split into a grid, according to discretizations of these features, to group similar solutions into one set.
MAP-Elites uses an evolutionary algorithm to produce and evolve new solutions, which are tested for their performance and to retrieve the values for their behavioural features. MAP-Elites keeps track of the best solution of each cell (elite) and, over multiple generations, produces high-performing solutions corresponding to different locations in the grid. Implicitly, MAP-Elites performs a quality-diversity optimization, as the distinction of cells enforces diversity of solutions and keeping the elites guarantees storing the best individuals found for each cell. This is especially interesting for the generation of agents with different play-styles (e.g. evolved vectors could represent the agent's preferred army composition, which will naturally yield different fighting styles).

 
\section{Game Playing Agent} \label{sec:agent}



\subsection{Scripts} \label{ssec:scripts}

Portfolio methods have a collection of scripts $\sigma_i \in \Sigma$. In Tribes, for a player $p$, each actor (i.e. the tribe, or any of its cities or units) $\lambda_i \in \Lambda$ has a collection of actions available at a particular decision-making step. The set of all available actions $A$ at a given step can be seen as the sum of the actions of all the entities that can still make a move in that turn.

In this work, scripts are closely tied to the action type they are used for. We employ $3$ different types of scripts attending to the nature of the action: \textit{wrapper}, \textit{best-choice} and \textit{style-choice} scripts. \textit{Wrapper} scripts are the simplest, returning the only possible action that can be executed for their actor. In other words, there is only one way of executing (or not) that action. Examples of these are Capture (a unit can only capture an enemy or neutral city if it is located at the same tile as the city is), Upgrade (a boat to a ship) or Ending the turn.

In contrast, all other actions have a \textit{choice}. For instance, a unit may have multiple places to move to, or a city may have multiple locations to construct a building. For these cases, the script can be a \textit{best-choice} or a \textit{style-choice} script. While the former scripts pick the best action possible among the possible options available for that actor, the latter offer different possibilities according to distinct ways of playing the game.

Each script receives a current game state and the set of available actions of the given action type (e.g. all possible target units for an attack) for the actor. The script returns the best action among the available ones and a value for that action, $v(a) \in [0,1]$ for a script $\sigma$, determined by domain knowledge. $v(a)$ is arbitrarily set to $0.5$ for all \textit{wrapper scripts}, given there is only one action available to choose from. 

\subsection{Portfolio MCTS Variants}\label{sec:pmcts}

\subsubsection{Portfolio MCTS (\textbf{P-MCTS})} \label{ssec:pmcts}

The modification required to turn the MCTS agent into a Portfolio MCTS (P-MCTS) is simple: in MCTS, the action that links one node to the next is one of the possible actions that can be executed in the original game state. A superset of the actions available is the cross product of all action types (see second column of Table~\ref{tab:tabScripts}) by all existing actors that have not completed their move in the current turn, considering only the valid actions. 

In P-MCTS, nodes are linked by \textit{action assignments}, which are tuples of the form $\tau = <\sigma, \lambda, a, v(a), w_\sigma>$: a script $\sigma$ is assigned to an actor $\lambda$, producing an action $a$ to execute, with a value $v(a) \in [0,1]$ as provided by the script. $w_\sigma$ is a weight associated with the script $\sigma$, which for the moment may be ignored (by default, $w_\sigma = 0.5$ $\forall \sigma \in \Sigma$). Action assignments are created in the Expansion phase of MCTS and the action that leads to the new state is automatically computed by the script and fixed for the rest of iterations of the algorithm. There is no differentiation for the opponent's turn with respect to processing the action scripts: all are computed assuming the current player is the max player. Then, the selection step in MCTS inverts the UCT values as in default 2-player MCTS. 

Table~\ref{tab:tabScripts} lists the $57$ scripts used in this study. This high number of scripts, as shown in the results (Section~\ref{sec:res}), permits the design of a competitive agent while allowing for diverse play-styles. \textit{Wrapper} and \textit{best-choice} scripts filter out weak actions, generally providing a stronger level of play, while \textit{style-choice} scripts not only eliminate poor choices, but also permit different ways of carrying out \textit{good} choices.

\subsubsection{P-MCTS with Progressive Bias (\textbf{P-MCTS (B)})}

Progressive Bias~\cite{chaslot2008} is an MCTS enhancement that modifies the selection step of the algorithm to add domain knowledge in order to prioritize actions preferred by a heuristic. The default UCB policy is modified to be computed as in Equation~\ref{eq:ucb1pb}.

\begin{equation}
    \argmax_{a \in A} \quad  Q(s,a) + C \sqrt \frac{ln N(s)}{N(s,a)} + \phi 
\label{eq:ucb1pb}
\end{equation}

The first two terms are the classical components of UCB: $Q(s,a)$ is the exploitation term, estimate of how good action $a$ is for state $s$ based on the rewards observed. The second expression is the exploration term, which gives more weight to states that have been explored less. $N(s)$ is the number of visits of the state $s$ and $N(s,a)$ indicates the number of times $a$ has been selected from $s$. The constant $C$ balances between these two terms. Progressive bias adds the third term, $\phi = \frac{h(s,a)}{1 + N(s,a)} $, defining a heuristic expression $h(s,a)$ for action $a$ in state $s$, which is inspired by expert domain knowledge. 

In this work, the heuristic function for progressive bias is associated with the action assignment, and it is defined as in Equation~\ref{eq:pb}. The heuristic value of a move in a state is determined by the value given by the script, multiplied by the weight assigned to that script.

\begin{equation}
    \phi = \frac{h(\tau)}{1 + N(s,a)} = \frac{v(a) \times  w_\sigma}{1 + N(s,a)}
\label{eq:pb}
\end{equation}

\subsubsection{P-MCTS with Progressive Unpruning (\textbf{P-MCTS (PU)})} 

Progressive Unpruning (PU;~\cite{chaslot2008}) is an MCTS enhancement especially designed for problems with high branching factors. PU first reduces the branching factor artificially for every node for the selection step of MCTS: When $N(s) > T$, it prunes all nodes except the $k_{0}$ nodes with the highest value $h(\cdot)$. Then, as the number of iterations through this node increases, the best pruned node as indicated by $h(\cdot)$, is unpruned according to the schedule $\beta_2\beta_1^{k - k_{0}}$ (where $k$ is the $k^{th}$ node to be unpruned).  The original PU algorithm uses constants for the parameters $T$, $\beta_1$, $\beta_2$ and $k_0$. This may be appropriate for problems with an uniform action space size, but tuning these in a turn-based strategy game which such a variable action space $|A(s)|$ is not trivial. Thus, we opted for a version that can dynamically adapt to the size of the changing action space: for P-MCTS, $k_{0} = |A(s)| \times \alpha_k$, $T = |A(s)| \times \alpha_t$ and $\beta_1 = |A(s)| \times \alpha_{\beta}$.
We tuned  $\alpha_k = 0.5$, $\alpha_t = 2.0$, $\alpha_{\beta} = 3.0$ and $\beta_2 = 1.3$ to provide a flexible pruning schedule that depends on $|A(s)|$.





\section{MAP-Elites for Exploring Competitive Play-Styles} \label{sec:mapelites}

Our implementation of MAP-Elites maps from the vector space $\mathcal{W} = \{w_{\sigma_1}, w_{\sigma_2}, \dots w_{\sigma_n}\}$ to a two-dimensional feature space. Each script has an associated weight $w_\sigma$ which, by default, is set to $0.5$. MAP-Elites evolves vectors within the space of weights $\mathcal{W}$, which in turn determine how P-MCTS (PU) prioritizes scripts to execute, prune and unprune. 
A log is kept that records gameplay statistics to be used for inferring the MAP-Elites dimensions, including (in total and per turn) score, actions executed, number of units, buildings and cities, tiles owned, production and technologies researched.

Algorithm~\ref{alg:mapelites} shows the pseudocode of MAP-Elites for generating diverse players with P-MCTS (PU). This pseudocode shows two functions: \textit{EvaluateIndividual} and \textit{MAP-Elites}. \textit{EvaluateIndividual} (lines~\ref{alg:line:evalStart} to~\ref{alg:line:evalEnd}) receives a set of weights that are used to initialize the player $P$ (line~\ref{alg:line:initP}), which is then used to play $|L| \times R_L$ games, where $L$ is a set of training levels and $R_L$ the number of repetitions each level is played. Gameplay statistics are accumulated for all these games (line~\ref{alg:line:accum}), which are then used to extract the desired phenotypical features for MAP-Elites (line~\ref{alg:line:feat}). These features $\phi_1, \phi_2$ are used to identify the cell in the MAP-Elites grid that this individual belongs to. At most $1$ individual is kept in each cell: if the evaluated individual is better than the one currently in the grid (or if there's none), it enters the map in that position (lines~\ref{alg:line:ifBetter} and~\ref{alg:line:evalEnd}). In our design, an individual is better than another if it achieves a higher win rate over the $|L| \times R_L$ games played, using the highest average score as a tie breaker.

\begin{algorithm}[!t]
\hrulefill

 \textit{Input:} $L$: set of levels $\{ l_1, l_2, \dots, l_m \}$. \newline  
 \textit{Input:} $R_L$: level repetitions. \newline 
 \textit{Input:} $P$: P-MCTS (PU) player. \newline
 \textit{Input:} $R_M$: MAP-Elites Random Initializations. \newline
 \textit{Input:} $I_M$: MAP-Elites Number of iterations. \newline
 \textit{Input:} $N_W$: Number of weights to evolve. \newline
 \textit{Input:} $\Phi$: Gameplay features for MAP-Elites. \newline
 \textit{Output:} $MAP$: final map of elite individuals.

\hrulefill

\begin{algorithmic}[1]
\Function{EvaluateIndividual}{$\mathcal{W}$}   \label{alg:line:evalStart}
    \State $GameplayStats \leftarrow \emptyset$ 
    \State $P$.init($\mathcal{W}$) \algorithmiccomment{Init P-MCTS (PU) w/ weights $\mathcal{W}$} \label{alg:line:initP}
    \For {level $l_i$ in $L$} \algorithmiccomment{play $L$ levels}
        \For {all reps in $R_L$} \algorithmiccomment{$R_L$ times each}
            \State $log$ \textleftarrow\; PlayGame($l_i$)
            \State $GameplayStats$.Add ($log$)  \label{alg:line:accum}
        \EndFor 
    \EndFor 
    \State $\phi_1, \phi_2$ \textleftarrow\; GameplayStats.ExtractFeatures($\Phi$) \label{alg:line:feat}
    \If {$MAP[\phi_1, \phi_2]$.FoundBetter($GameplayStats$)} \label{alg:line:ifBetter}
        \State $MAP[\phi_1, \phi_2] \leftarrow \mathcal{W}$ \algorithmiccomment{Substitute elite if better} \label{alg:line:evalEnd}
    \EndIf
\EndFunction    
\Function{MAP-Elites}{} \label{alg:line:mapStart}
    \State $MAP \leftarrow \emptyset$ 
    \For {weight $w_i$ in $N_W$} \algorithmiccomment{Initial Mapping} \label{alg:line:mappingStart}
        \State $\mathcal{W} \leftarrow Array(0.0)$  \algorithmiccomment{Set of weights, all 0.0}
        \State $\mathcal{W}[w_i] \leftarrow 1.0$    \algorithmiccomment{Only one weight set at 1.0}
        \State $\call{EvaluateIndividual}{(\mathcal{W})}$ \label{alg:line:mappingEnd}
    \EndFor 
    
    \For {iteration $i$ in $R_M$} \algorithmiccomment{Random Initialization} \label{alg:line:rndStart}
        \State $\mathcal{W} \leftarrow RandomArray(0.0, 0.1)$
        \State $\call{EvaluateIndividual}{(\mathcal{W})}$ \label{alg:line:rndEnd}
    \EndFor 
    
    \For {iteration $i$ in $I_M$} \algorithmiccomment{Main algorithm iterations}
        \State $\mathcal{W}_1 \leftarrow MAP.RandomElite()$ \label{alg:line:rndPick}
        \State $\mathcal{W}_2 \leftarrow \mathcal{W}_1.Mutate()$ \algorithmiccomment{Generate new weight set} \label{alg:line:rndMutate}
        \State $\call{EvaluateIndividual}{(\mathcal{W}_2)}$ \label{alg:line:evalInd}
    \EndFor 
    \State  \Return $MAP$ \label{alg:line:mapEnd}
\EndFunction
\end{algorithmic}
\caption{Pseudocude of the MAP-Elites algorithm for generating diverse Play-Styles with P-MCTS (PU).}\label{alg:mapelites}
\end{algorithm}

The function \textit{EvaluateIndividual} is used by \textit{Map-Elites} (lines~\ref{alg:line:mapStart} to~\ref{alg:line:mapEnd}) to populate a map with individuals. The process is simple and divided into three blocks. First, an initial mapping of the multidimensional input space is performed by evaluating individuals which have only one weight $w_{\sigma_i} = 1.0$, with $w_{\sigma_j} = 0.0$ $\forall i \neq j$ (lines~\ref{alg:line:mappingStart} to~\ref{alg:line:mappingEnd}). This is done for all weights $w_{\sigma_i}$ in the vector, aiming to have a diverse initialization of the grid. The second step is to evaluate $R_M$ random vectors (lines~\ref{alg:line:rndStart} to~\ref{alg:line:rndEnd}). Finally, the existing population of individuals is evolved, during $I_M$ iterations, generating new individuals mutating existing ones. In this work, we employ a simple Stochastic Hill Climber, which takes an individual at random from the map (line~\ref{alg:line:rndPick}), mutates one weight uniformly at random (line~\ref{alg:line:rndMutate}) and evaluates the resulting weight vector (line~\ref{alg:line:evalInd}).


\section{Experimental Work and Results} \label{sec:res}

\subsection{Weighted Progressive Unpruning} \label{ssec:res-pmcts}

Table~\ref{tab:resPMCTS} shows the performance of the different versions of Portfolio MCTS (P-MCTS, P-MCTS (B) and P-MCTS (PU)) versus other agents in the framework: default MCTS and RHEA. 
Each row shows the result of $500$ two-player games of Tribes indicating, from left to right: win rate, 
score, percentage of the technology tree researched, number of cities and production at the end of the game. The $500$ games are distributed among $25$ different levels played $20$ times each, with agents alternating positions to account for potentially unbalanced maps. These levels are the same as those used in~\cite{perez2020tribes}, to allow for a direct comparison. The parameters of the algorithms are also equivalent to those from~\cite{perez2020tribes}: play-out and individual length of all MCTS, Portfolio MCTS and RHEA agents is $20$; tree selection constant $C = \sqrt{2}$; rewards bounded in the [$0$, $1$] interval; decision budget per action decision-making is $2000$ usages of the forward model; population size of RHEA is $1$. States are valued by a linear combination of state features, computed as differential variables (production, researched technologies, cities owned, units, etc.) between starting and end states in a play-out or individual. See~\cite{perez2020tribes} for a full description of this state evaluation function and level seeds, as well as in the public repository\footnote{https://github.com/GAIGResearch/Tribes/tree/portfolio}.  
Results clearly show that the inclusion of the portfolio significantly improves the win rate with respect to MCTS. In direct comparison, P-MCTS beats MCTS $65.60\%$ of the time. When playing against RHEA, P-MCTS also obtains a higher win rate 
than MCTS, showing an improvement also in the other indicators shown in the table. In particular, the gain in win rate against RHEA is very substantial with the portfolio. Despite the high number of scripts, the branching factor is reduced considerably, because $n$ actions for an actor and action type are substituted with one single script. The average turn branching factor in P-MCTS is $10^3$ for early and $10^{10}$ for end game, which constitutes a reduction of several orders of magnitude compared to MCTS.

Progressive bias alone does not show an improvement in win rate, but the addition of (un)pruning does improve this indicator across the board. Table~\ref{tab:resPMCTS} shows an interesting consequence of progressive unpruning in MCTS against all agents: while many indicators such as the score, percentage of technologies and the cities owned at the end are lower, win rate increases. This shows the effectiveness of P-MCTS (PU): to win, a player does not need to achieve the highest production or research all technologies - only to capture the opponent's capital.




\begin{table}[!t]
    \caption{Performance of variants of Portfolio MCTS and Pruning methods. Statistics for all games are averaged across $500$ games. Values between brackets indicate standard error. }
    \label{tab:resPMCTS}
    \centering
    \begin{tabular}{|l|>{\centering\arraybackslash} m{0.75cm}|>{\centering\arraybackslash}  m{0.85cm}|>{\centering\arraybackslash} m{0.75cm}|>{\centering\arraybackslash} m{0.55cm}|>{\centering\arraybackslash} m{1.25cm}|}
\hline

\textbf{Agent} & \textbf{Wins} & \textbf{Score} & \textbf{Techs} & \textbf{Cities} & \textbf{Production}\\
\hline
\multicolumn{6}{|c|}{\textbf{vs MCTS}} \\\hline
\textbf{P-MCTS} & 65.60\% (2.93) & 7237.36 (323.66) & 93.02\% (4.16) & 3.01 (0.13) & 18.25 (0.82) \\
\hline
\textbf{P-MCTS (B)} & 65.40\% (2.92) & 7290.61 (326.05) & 92.94\% (4.16) & 3.03 (0.14) & 18.92 (0.85) \\
\hline
\textbf{P-MCTS (PU)} & 68.60\% (3.07) & 6330.35 (283.10) & 83.29\% (3.72) & 3.23 (0.14) & 16.95 (0.76) \\
\hline
\multicolumn{6}{|c|}{\textbf{vs RHEA}} \\\hline
\textbf{MCTS}  & 37.00\% (2.16) & 4274.10 (191.14) & 84.17\% (3.76) & 1.69 (0.08) & 13.13 (0.59) \\
\hline
\textbf{P-MCTS} & 56.40\% (2.52) & 6734.93 (301.20) & 89.38\% (4.00) & 2.73 (0.12) & 15.62 (0.70) \\
\hline
\textbf{P-MCTS (B)} & 55.60\% (2.49) & 6745.07 (301.65) & 90.11\% (4.03) & 2.66 (0.12) & 15.19 (0.68) \\
\hline
\textbf{P-MCTS (PU)} & 67.80\% (3.03) & 5916.11 (264.58) & 77.01\% (3.44) & 3.17 (0.14) & 15.66 (0.70) \\

\hline
    \end{tabular}
    
\end{table}

\subsection{Generating Diverse Play-styles}


The following set of experiments shows how we use MAP-Elites to generate diverse play-styles, using the algorithm explained in Section~\ref{sec:mapelites}. All games run for this experimental setup are played between a P-MCTS (PU) player and the default MCTS player, using the same agent parameterizations described above. The evaluation of an individual for MAP-Elites can be computationally expensive if different levels and a high number of repetitions are used. Running $500$ per individual, as done for the tests described in Section~\ref{ssec:res-pmcts}, is unfeasible. Thus, we select a reduced set of levels to run MAP-Elites evaluations, but picking those for which the win-rate observed in the previous experiments between P-MCTS (PU) and MCTS is closer to the overall win-rate across the $25$ levels. We pick the top $5$ levels, which are played $4$ times each, resulting in $|L| \times R_L = 5 \times 4 = 20$ games played per individual.

For this work, not all weights of the space $\mathcal{W}$ are evolved, as this would create individuals of length $57$, making this evolutionary process very slow. Instead, we only evolved the weights corresponding to the Research and Spawn action types ($\sigma_{15}$ to $\sigma_{19}$ and $\sigma_{36}$ to $\sigma_{41}$), which can substantially influence the player's strategy. The rest of the weights are set to their default value of $0.5$. Picking these weights to evolve gives an individual length of $N_W = 11$. Each weight can have $5$ different values: [$0.0$, $0.25$, $0.5$, $0.75$, $1.0$], which opens a search space of $5^{11} \approx 5 \times 10^7$ points. With this space we aim to show that, even in relatively not very large search spaces, we are able to obtain a varied set of different play-styles. We hypothesize that a more diverse set of behaviours can be achieved with a higher search space, at the expense of a higher computational expense. For all our use-cases, we  randomly initialize the map with $R_M = 40$ individuals and perform a total of $I_M = 500$ iterations of MAP-Elites. 

We have run three different use-cases to illustrate the results that can be achieved with our proposed approach, which corresponds to different pairs of features (the two dimensions $\Phi$ of MAP-Elites) extracted from game-play data and used to assign the individuals to a cell in the map.

\begin{table}[!t]
    \caption{Features of individuals obtained with MAP-Elites (col. 3) and validation process (col. 4). Each row represents one individual, its use case (I, II, III) and their cell in the map.}
    \label{tab:resAgg}
    
    \centering
    \begin{tabular}{|>{\centering\arraybackslash} m{1.45cm}|>{\centering\arraybackslash} m{2.8cm}|>{\centering\arraybackslash} m{1.45cm}|>{\centering\arraybackslash} m{1.25cm}|}
\hline

\textbf{MAP-Elites Cell} & \textbf{Feature} & \textbf{MAP-Elites Result} & \textbf{Validation} \\
\hline

\multirow{3}{*}{I: [14:4]} & Win Rate & 60\% & 55.2\% \\
 & $\phi_1:$ \# Attacks & 12.6 & 13.83 \\
 & $\phi_2:$ Support Unit & 3.65 & 3.792 \\
\hline

\multirow{3}{*}{I: [6:-1]} & Win Rate & 75\% & 70.2\% \\
 & $\phi_1:$ \# Attacks & 5.4 & 9.7 \\
 & $\phi_2:$ Support Unit & -0.75 & -0.726 \\
\hline
\hline

\multirow{3}{*}{II: [0:25]} & Win Rate & 75\% & 75.2\% \\
 & $\phi_1:$ Defender Spawns & 0.4 & 0.8 \\
 & $\phi_2:$ Win Turn & 23.1 & 22.9 \\
\hline

\multirow{3}{*}{II: [7:40]} & Win Rate & 60\% & 62\% \\
 & $\phi_1:$ Defender Spawns & 6.5 & 4.6 \\
 & $\phi_2:$ Win Turn & 39.5 & 21.6 \\
\hline
\hline

\multirow{3}{*}{III: [0.6:0.6]} & Win Rate & 50\%  & 58.1 \%  \\
 & $\phi_1:$ Research Progress & 0.63 & 0.43 \\
 & $\phi_2:$ Tile Dominance & 0.664 & 0.44 \\
\hline

\multirow{3}{*}{III: [1.2:1.0]} & Win Rate & 85\% & 74.37 \% \\
 & $\phi_1:$ Research Progress & 1.05 & 0.7 \\
 & $\phi_2:$ Tile Dominance & 0.87 & 0.65 \\
\hline

    \end{tabular}
    
\end{table}

\begin{figure*} [!t]
\centering
\includegraphics[width=0.99\textwidth]{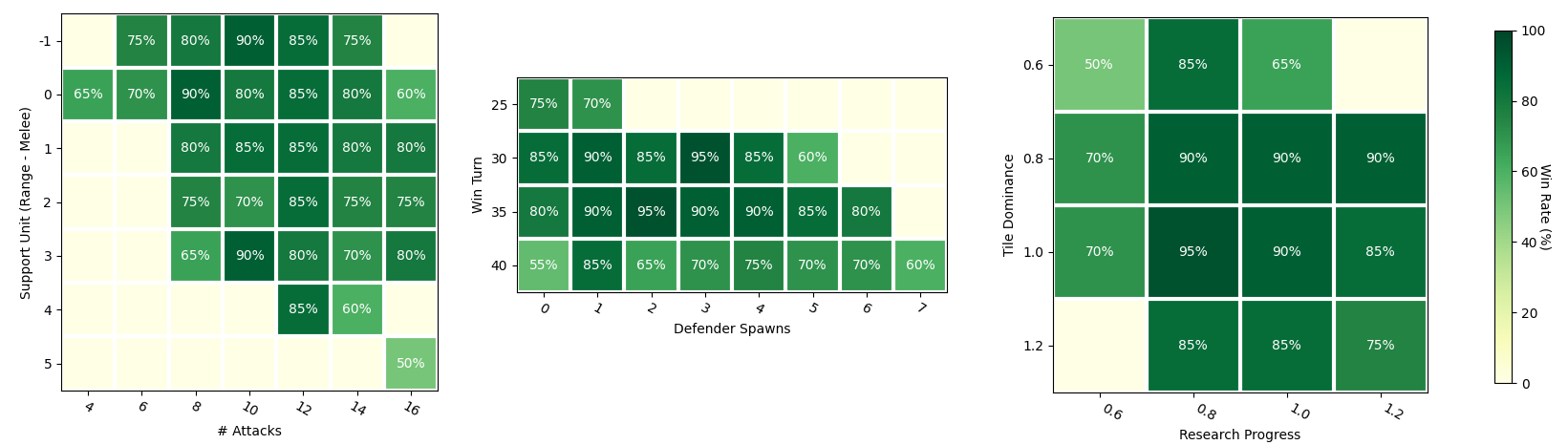}
\caption{Final maps as evolved by MAP-Elites for use-cases I (left), II (centre) and III (right). Axis are MAP-Elites features and the percentage indicated in each elite (cell) is the win rate of the P-MCTS (PU) agent versus MCTS. These plots only show the inhabited portion of the MAP-Elites grids.} 
\label{fig:allgrids}
\end{figure*}

\begin{figure*} [!t]
\centering
\includegraphics[width=0.99\textwidth]{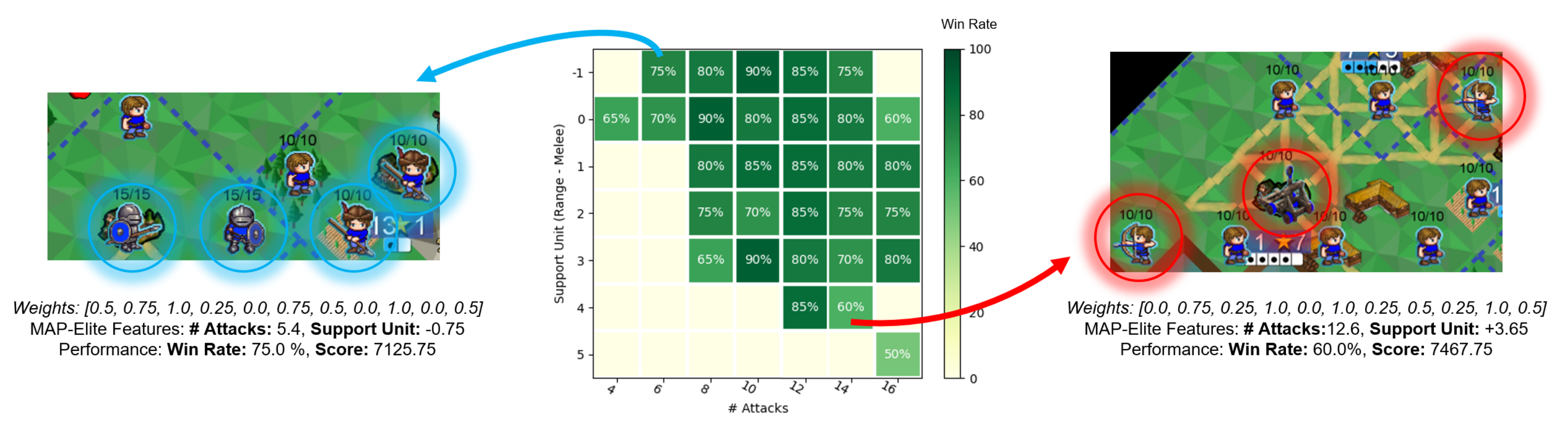}
\caption{Example and screenshots of two games from the map evolved for use-case I. Circled units highlight melee support (left) and ranged support (right) units. Weights (in the order $\sigma_{15}$ to $\sigma_{19}$ and $\sigma_{36}$ to $\sigma_{41}$ as per Table~\ref{tab:tabScripts}), features and performance is also indicated for each one of these elites.} 
\label{fig:mapExample}
\end{figure*}

\subsubsection*{I) \# Attacks vs Support Units} In this use-case we map the number of attack actions executed ($\phi_1$) and the type of support units ($\phi_2$) used by P-MCTS (PU). For the latter, we define two groups of units that complement \textit{Warrior} units, which form the backbone of the army. One group is melee ($M$: formed by \textit{Knights} and \textit{Swordsmen}) and the other is ranged ($R$: \textit{Archers} and \textit{Catapults}). This feature corresponds to the difference of spawn units between these two groups, so that $\phi_2 = |R|-|M|$. A positive value shows the player relies more on ranged support units, while a negative number generates more melee units. The final map can be seen in Figure~\ref{fig:allgrids} (left). The coloured cells indicate positions in the final map that hold an individual, and the indicated value corresponds to the win rate of the P-MCTS (PU) agent. The map uses $\phi_1 \in [0, 20]$ with increments of $2$, and $\phi_2 \in [-5,5]$ with increments of $1$, which resulted in $30$ different individuals. Note that the evolved win rates may be higher than the one shown in Table~\ref{tab:resPMCTS}: this is a direct consequence of MAP-Elites keeping the best individual ever found for each cell and also optimizing against a fixed agent. Figure~\ref{fig:mapExample} shows an example of two individuals obtained by this run, including screenshots of the games, evolved weights, MAP-Elite feature values and individual performance. In order to validate that the individuals evolved by MAP-Elites do showcase different characteristics of game-play, we have run an extensive experimentation of $500$ games, same as in Section~\ref{ssec:res-pmcts}, to confirm that the expressed behaviour extends to the rest of the levels of the game. Table~\ref{tab:resAgg} compares the performance of the MAP-Elites and the validation runs for two individuals of each use-case, corresponding to opposite corners of the resulting map. As can be seen, there is a clear correlation of values for both features (first two rows) between the evolution and the validation results.

\subsubsection*{II) Defender Spawns vs Winning Turn} In our second use case, we record the number of spawns of the \textit{Defender} unit type ($\phi_1$) and the average turn at which P-MCTS (UP) achieves victory ($\phi_2$). The rationale behind this is to explore different play-styles with respect to having a more or less Defender units and the ability to win the game after fewer or more turns. The resultant map is shown in Figure~\ref{fig:allgrids} (centre), eliciting $23$ different individuals. It's worth observing the distribution of these configurations in the map, showing a lack of individuals in the upper right corner. This result is sensible: that part of the map corresponds to winning the game quickly ($35$ turns or less) while spawning many defensive units ($\geq 5$). The more defensive the strategy is, the harder is to win the game quickly. In any case, the agents obtained are still able to keep a high win rate \textit{independently} of this strategy.

Rows $3$ and $4$ of Table~\ref{tab:resAgg} show the validation results for two individuals of this map. Although the features again present similar values, it's worth mentioning that ($\phi_2$) for the individual in row $4$ achieves victories much quickly in validation than in the results from MAP-Elites. Our interpretation of this divergence is that the feature \textit{Win Turn} is intrinsically related to the win rate: the number of sample points for this average is lower than for other features (only $60\%$ in this case: $12$ of the $20$ games played). This increases the variance of the measure and may lead to inaccurate evaluations. One solution could be to increase the number of evaluations where this situation can happen, or to redefine how elites are kept considering their fitness variance. Another one, which inspired the following use-case, is to look for other ways of representing progress rate towards victory that do not depend on the winning condition.    

\subsubsection*{III) Research Progress vs Tile Dominance} This final use-case compares features that relate to how fast progress is made towards victory. Two aspects are considered: research progress ($\phi_1$) and tile dominance ($\phi_2$). Their values are computed as the the slope of the function that represents the number of technologies researched (for $\phi_1$) and board tiles owned (for $\phi_2$) per turn, calculated with a linear regression. These features will have a value of $1.0$ if the P-MCTS (PU) agent researches one technology (or acquires control of a new tile, respectively) per turn. Figure~\ref{fig:allgrids} (right) shows the map obtained for this run, with $14$ individuals distributed across the cells. Not surprisingly, win rates are higher when both features are also higher. This is corroborated in the validation shown in the last two rows of Table~\ref{tab:resAgg}, which also shows that the values for these features consistently present a higher progression for dominance and research according to their respective cells. This example also reveals an interesting application of this method: setting opponent AI difficulty levels by picking individuals from different cells in the evolved grid.


\section{Conclusions and Future Work} \label{sec:end}

This paper presents a variation of MCTS that uses a portfolio of scripts and Progressive Unpruning (PU) to tackle large action spaces in turn-based strategy games. Rather than focusing only on the playing strength of the final agent, we explore how different yet still competitive behaviours can be obtained from the portfolio MCTS agent using MAP-Elites. Results show two clear outcomes: first, that our implementation of Portfolio MCTS with PU clearly outperforms the MCTS agent in this game. Section~\ref{ssec:res-pmcts} and Table~\ref{tab:resPMCTS} show that the improvement comes both from i) using a portfolio of scripts to reduce the action space; and ii) pruning, which improves win rate when compared to non-pruned Portfolio MCTS. 
Secondly, our results show that it's possible to parameterize Portfolio MCTS to bias pruning in order to achieve different play-styles that are still competitive. Our use-cases show that MAP-Elites allows to explore and differentiate resulting behaviours, which are mapped to different pairs of gameplay features. The $3$ use-cases shown in this paper are just some of the possible features that can be used for identifying different play-styles, but there is a wider space of possibilities to alter these. Not only other features can be constructed according to different needs, but also other weights could be evolved in order to achieve diverse behaviours. The examples presented here show two-dimensional MAP-Elites, but it is also possible to map behaviours across more features. Preliminary tests (not included here due to space limitations) show that diverse behaviours can also be obtained with three behavioural features.
Finally, our work shows the importance of identifying which features may capture a desired behaviour, and how they may affect the validation of the evolved behaviours in a larger setting.

We identify two immediate lines of future work. First, to investigate if other quality-diversity algorithms (like constrained MAP-Elites~\cite{khalifa2018talakat}) are able to expand the diversity of achievable behaviours, or to maximize the occupancy of the final map with individuals that accurately reflect specific game-play traits. Given the stochastic nature of these evaluations, one possibility is to substitute our Stochastic Hill Climber for a method suited to deal with noisy environments, like the N-Tuple Bandit Evolutionary Algorithm~\cite{lucas2019efficient}. Another interesting line of work is to explore the influence of the evolved weights for the scripts on the final behaviours. Not only to infer which scripts are more relevant for which play-styles (which may be quite specific to each game), but also to devise mechanisms to automatically estimate the sensitivity of the different weights and which ones are more relevant to achieve which behaviours. This could have interesting consequences for game designers, as it would open the possibility of identifying which scripts (and therefore actions) have a greater impact on the game-playing strategies.


\begin{table*}[!t]
        \caption{Scripts for Tribes, grouped by type. From left to right: script type, action type (T: tribe action, C: city action, U: unit action), play-style script (for style-choice scripts only) and script id plus brief description.}
    \label{tab:tabScripts}
    \centering
    \begin{tabular}{|c|c|c|l|}
\hline

\textbf{Script Type} & \textbf{Action Type} & \multicolumn{2}{c|}{\textbf{Script $\sigma_{i}$: Description}} \\
\hline
\multirow{10}{*}{Wrapper} 
 & End Turn (T)  & \multicolumn{2}{l|}{ $\sigma_{1}$: Ends the current turn. } \\
\cline{2-4}
 & Destroy (C)  & \multicolumn{2}{l|}{ $\sigma_2$: Destroys a target building owned by the tribe. } \\
\cline{2-4}
 & Examine  (U) & \multicolumn{2}{l|}{ $\sigma_3$: Explores ruins for a bonus. } \\
\cline{2-4}
 & Heal Others  (U) & \multicolumn{2}{l|}{ $\sigma_4$: Heal other units around this one (Mind Bender only).} \\
\cline{2-4}
 & Recover (U)  & \multicolumn{2}{l|}{ $\sigma_5$: Recover hit points and finish movement. } \\
\cline{2-4}
 & Upgrade Boat (U)  & \multicolumn{2}{l|}{ $\sigma_6$: Upgrades a boat to a ship. } \\
\cline{2-4}
 & Upgrade Ship (U)  & \multicolumn{2}{l|}{ $\sigma_7$: Upgrades a ship to a battleship. } \\
\cline{2-4}
 & Make Veteran (U)  & \multicolumn{2}{l|}{ $\sigma_8$: Makes this unit a veteran (after killing three units). } \\
\cline{2-4}
 & Capture (U) & \multicolumn{2}{l|}{ $\sigma_9$: Captures an enemy city or a neutral village. } \\
\cline{2-4}
 & Disband (U)  & \multicolumn{2}{l|}{ $\sigma_{10}$: Disbands a target unit owned by the tribe. } \\
\hline

\multirow{4}{*}{Best-choice} & Build Road (T) & \multicolumn{2}{l|}{ $\sigma_{11}$: Builds a road in a non-enemy controlled tile. } \\
\cline{2-4}
 & Burn Forest (C)  & \multicolumn{2}{l|}{ $\sigma_{12}$: Burns a forest to create crop resource. } \\
\cline{2-4}
 & Grow Forest (C) & \multicolumn{2}{l|}{ $\sigma_{13}$: Grows a Forest in a tile. } \\
\cline{2-4}
 & Get Resource (C) & \multicolumn{2}{l|}{ $\sigma_{14}$:  Gathers a resource from a tile within the city's borders. } \\
\hline

\textbf{Script Type} & \textbf{Action Type} & \textbf{Play-Style Script} & \multicolumn{1}{c|}{\textbf{Script $\sigma_{i}$: Description}} \\
\hline

\multirow{40}{*}{Style-choice} & \multirow{5}{*}{Research (T)}  & Farms & $\sigma_{15}$:  Researches the lowest tier technology possible in the Farms branch of the tech tree. \\
\cline{3-4}
 &  & Naval & $\sigma_{16}$: Researches the lowest tier technology possible in the Naval branch of the tech tree. \\
\cline{3-4}
 &  & Mountain & $\sigma_{17}$: Researches the lowest tier technology possible in the Mountain branch of the tech tree. \\
\cline{3-4}
 &  & Range & $\sigma_{18}$: Researches the lowest tier technology possible in the Range branch of the tech tree. \\
\cline{3-4}
 &  & Roads & $\sigma_{19}$:  Researches the lowest tier technology possible in the Network branch of the tech tree. \\
 \cline{2-4}

 & \multirow{4}{*}{Clear Forest (C)}  & For Custom House & $\sigma_{20}$: Removes a forest to create a spot for a custom house. \\
\cline{3-4}
 & & For Forge & $\sigma_{21}$: Removes a forest to create a spot for a forge. \\
\cline{3-4}
 & & For Sawmill & $\sigma_{22}$: Removes a forest to create a spot for a sawmill. \\
\cline{3-4}
 & & For Windmill & $\sigma_{23}$: Removes a forest to create a spot for a windmill. \\
 \cline{2-4}
 
 & \multirow{2}{*}{Level Up  (C)} & Growth & $\sigma_{24}$: Levels a city up and chooses a bonus that maximizes city production. \\
\cline{3-4}
 &  & Military & $\sigma_{25}$: Levels a city up and chooses a bonus that maximizes combat strength. \\
 \cline{2-4}
 
 & \multirow{10}{*}{Build (C)} & Custom House & $\sigma_{26}$: Builds a custom house in a tile with maximum neighbouring ports capacity. \\
\cline{3-4}
 &  & Windmill & $\sigma_{27}$: Builds a windmill in a tile with maximum neighbouring farms capacity.  \\
\cline{3-4}
 &  & Sawmill & $\sigma_{28}$: Builds a custom house in a tile with maximum neighbouring forest capacity. \\
\cline{3-4}
 &  & Forge & $\sigma_{29}$: Builds a custom house in a tile with maximum neighbouring ore mountains capacity. \\
\cline{3-4}
 &  & Port & $\sigma_{30}$: Builds a farm in a water tile, prioritizing those with neighbouring custom houses.\\
\cline{3-4}
 &  & Farm & $\sigma_{31}$: Builds a farm in a tile, prioritizing those with neighbouring windmills. \\
\cline{3-4}
 &  & Lumber Hut & $\sigma_{32}$: Builds a lumber hut in a tile, prioritizing those with neighbouring sawmills. \\
\cline{3-4}
 &  & Mine & $\sigma_{33}$: Builds a mine in a mountain tile with ore, prioritizing those with neighbouring forges. \\
\cline{3-4}
 &  & Monument & $\sigma_{34}$: Builds a monument in a tile owned by the city.\\
\cline{3-4}
 &  & Temple & $\sigma_{35}$: Builds a temple in a tile owned by the city.\\
 \cline{2-4}
  & \multirow{6}{*}{Spawn (C)} & Strongest & $\sigma_{36}$: Spawns the strongest available unit. \\
\cline{3-4}
 &  & Defensive & $\sigma_{37}$: Spawns the unit with the highest defensive value. \\
\cline{3-4}
 &  & Cheapest & $\sigma_{38}$: Spawns the cheapest available unit. \\
\cline{3-4}
 &  & Fastest & $\sigma_{39}$: Spawns the unit with the highest movement value. \\
\cline{3-4}
 &  & Highest HP & $\sigma_{40}$: Spawns the unit with the highest hit points. \\
\cline{3-4}
 &  & Range & $\sigma_{41}$: Spawns the unit with the highest range attack value. \\
 \cline{2-4}
 
 & \multirow{3}{*}{Convert (U)}  &  Strongest unit & $\sigma_{42}$: Converts to the current tribe the strongest enemy unit in range. \\
\cline{3-4}
 &  &  Highest HP & $\sigma_{43}$: Converts to the current tribe the enemy unit in range with highest hit points. \\
\cline{3-4}
 &  & Highest Defence & $\sigma_{44}$: Converts to the current tribe the enemy unit with the highest defence value. \\
 \cline{2-4}
 
 & \multirow{10}{*}{Move (U)} & To capture & $\sigma_{45}$: Moves the unit into an enemy city or neutral village. \\
\cline{3-4}
 & & To city centre & $\sigma_{46}$: Moves the unit to the tile of the closest owned city. \\
\cline{3-4}
 &  & Defensively & $\sigma_{47}$: Moves the unit towards the closest owned city. \\
\cline{3-4}
 &  & Offensively & $\sigma_{48}$: Moves the unit towards the closest enemy city. \\
\cline{3-4}
 &  & To Land & $\sigma_{49}$: Moves the unit to a land tile from a water one, disembarking. \\
\cline{3-4}
 &  & To Embark & $\sigma_{50}$: Moves the unit to a friendly port, embarking. \\
\cline{3-4}
 &  & To Attack Range & $\sigma_{51}$: Moves the unit to a position from which it can attack most enemy units.\\
\cline{3-4}
 &  & To Converge & $\sigma_{52}$: Moves the unit closer to other friendly units. \\
\cline{3-4}
 &  & To Diverge & $\sigma_{53}$: Moves the unit away from other friendly units. \\
 \cline{2-4}

 & \multirow{4}{*}{Attack (U)} & Closest & $\sigma_{54}$: Attacks the closest enemy unit in range. \\
\cline{3-4}
 &  & Weakest & $\sigma_{55}$: Attacks the enemy unit in range with the lowest defence value. \\
\cline{3-4}
 &  & Most Damaged & $\sigma_{56}$: Attacks the enemy unit in range with the lowest hit points remaining. \\
\cline{3-4}
 &  & Strongest & $\sigma_{57}$: Attacks the strongest enemy unit in range. \\
\hline
    \end{tabular}

\end{table*}

\section*{Acknowledgments}
Work supported by UK EPSRC grants EP/T008962/1, EP/L015846/1 and Queen Mary's Apocrita HPC facility. 

\bibliographystyle{IEEEtran}
\bibliography{IEEEabrv,main}

\end{document}